\documentclass[10pt,twocolumn,letterpaper]{article}

\usepackage{cvpr}              

\definecolor{cvprblue}{rgb}{0.21,0.49,0.74}
\usepackage[pagebackref,breaklinks,colorlinks,allcolors=cvprblue]{hyperref}

\usepackage{color}
\usepackage{multirow}
\usepackage{graphicx}

\def\paperID{17669} 
\def\confName{CVPR}
\def\confYear{2025}

\title{TADFormer : Task-Adaptive Dynamic TransFormer 

for Efficient Multi-Task Learning}

\author{Seungmin Baek\thanks{\noindent These authors contributed equally.} , Soyul Lee\footnotemark[1] , Hayeon Jo, Hyesong Choi, Dongbo Min\thanks{Corresponding author.}\\
Ewha Womans University, Korea\\
\tt\small {
\{min100kim,230aig,johayeon,hyesong,dbmin\}@ewha.ac.kr
}
}

\begin{document}
\maketitle
\begin{abstract}

Transfer learning paradigm has driven substantial advancements in various vision tasks. However, as state-of-the-art models continue to grow, classical full fine-tuning often becomes computationally impractical, particularly in multi-task learning (MTL) setup where training complexity increases proportional to the number of tasks. Consequently, recent studies have explored Parameter-Efficient Fine-Tuning (PEFT) for MTL architectures. Despite some progress, these approaches still exhibit limitations in capturing fine-grained, task-specific features that are crucial to MTL. In this paper, we introduce {\bf T}ask-{\bf A}daptive {\bf D}ynamic trans{\bf Former}, termed {\bf TADFormer}, a novel PEFT framework that performs task-aware feature adaptation in the fine-grained manner by dynamically considering task-specific input contexts. TADFormer proposes the parameter-efficient prompting for task adaptation and the Dynamic Task Filter (DTF) to capture task information conditioned on input contexts. Experiments on the PASCAL-Context benchmark demonstrate that the proposed method achieves higher accuracy in dense scene understanding tasks, while reducing the number of trainable parameters by up to 8.4 times when compared to full fine-tuning of MTL models. TADFormer also demonstrates superior parameter efficiency and accuracy compared to recent PEFT methods. 


\end{abstract}
    
\section{Introduction}
\label{sec:intro}

\begin{figure}[t]
  \centering
\includegraphics[width=1\linewidth]{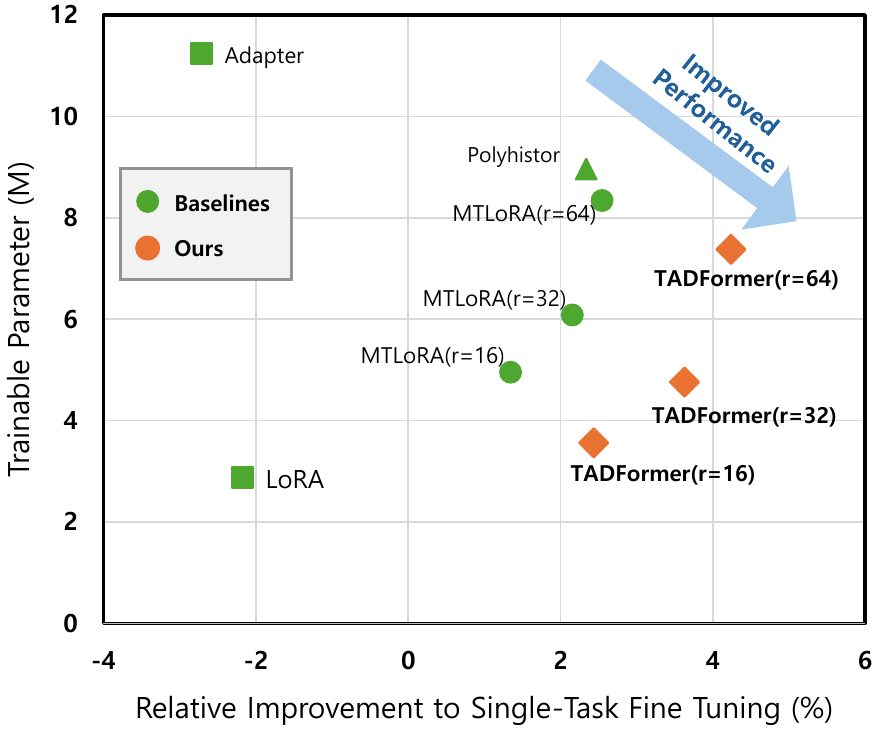}
\vspace{-20pt}
\caption{Performance comparison between TADFormer and baseline PEFT methods. Here, $r$ represents the rank of low-rank decomposition modules in both TADFormer and MTLoRA.}
\label{fig:figure1}
\end{figure}


Transfer learning paradigm has achieved significant advancements in the fields of natural language processing~\cite{devlin2018bert,brown2020language} and computer vision ~\cite{he2022mae,xie2022simmim}. Generally, given a pre-trained model on a large-scale dataset, the traditional transfer learning approaches fine-tune an entire model consisting of the pre-trained encoder and task-specific decoder for downstream tasks such as image classification, segmentation, and object detection. However, as the model capacity of state-of-the-art approaches continue to increase, applying the full fine-tuning to the entire model has become inefficient due to prohibitively high computational resource requirements.

To address this issue, recent works have begun to extensively explore Parameter-Efficient Fine-Tuning (PEFT)~\cite{hu2021lora,hu2023llmadapters,liu2022polyhistor,mahabadi2021parameter}, which aims to reduce the number of trainable parameters while maintaining the performance on the downstream tasks. For instance, adapter based approaches~\cite{houlsby2019parameter,hu2021lora} insert compact modules within Transformer blocks, visual prompt tuning~\cite{lester2021prompt,li2021prefix} leverages learnable tokens together with inputs or intermediate features, and Low-Rank Adaptation (LoRA)~\cite{hu2021lora} introduces trainable low-rank matrices into Transformer blocks for updating weights.

In a similar context, there have been active attempts to apply the PEFT to Multi-Task Learning (MTL) models~\cite{agiza2024mtlora,xin2024vmt}. MTL models~\cite{bhattacharjee2022mult,ye2023taskexpert,yang2024multi,huang2024synergy}, which aim to simultaneously process multiple tasks, generally require more training complexity proportional to the number of tasks, 
and the PEFT methodologies could be a promising solution to overcome this limitation. 
Multi-Task LoRA (MTLoRA)~\cite{agiza2024mtlora} extends LoRA~\cite{hu2021lora} to be more suitable for multi-task scenarios, while VMT-adapter~\cite{xin2024vmt} implements an MTL architecture based on adapters~\cite{houlsby2019parameter}. Although these methods facilitate parameter-efficient training within MTL models, they still exhibit inherent limitations in terms of 
leveraging the PEFT in the task-aware manner that fits the MTL architecture.


A primary challenge in training multiple tasks on the single model is to effectively leverage useful information across multiple tasks. 
To this end, MTL models should effectively capture the unique characteristics of each task while leveraging task-agnostic features by considering interactions between tasks~\cite{huang2024synergy,vandenhende2021multi}. 
According to existing MTL studies~\cite{ye2023taskexpert}, context information of input samples plays a critical role in enabling the model to capture the unique characteristics of each task. 
In other words, by considering the diversity of input samples during feature extraction, the model can learn finer-grained, task-specific characteristics. 
Moreover, when applying PEFT to MTL, only a subset of parameters is fine-tuned, unlike full-tuning, which can limit the model’s adaptability across diverse tasks. This necessitates the approach to extract task-specific features dynamically conditioned on input contexts.
However, current PEFT approaches for the MTL model~\cite{agiza2024mtlora,xin2024vmt} do not reflect sample dependency, extracting task characteristics using only static learnable parameters for each task, which limits their ability to capture task-specific features effectively with only very compact trainable modules.

\begin{figure}
  \centering
  \begin{subfigure}[b]{0.61\linewidth}
    \centering
    \includegraphics[width=\linewidth]{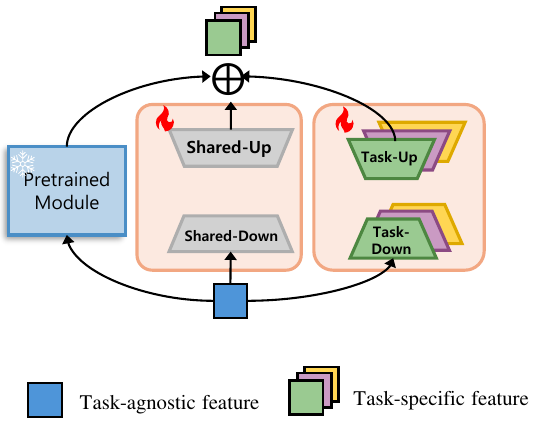}
    \caption{MTLoRA}
    \label{fig:figure2a}
  \end{subfigure}
  \hfill
  \begin{subfigure}[b]{0.38\linewidth}
    \centering
    \includegraphics[width=\linewidth]{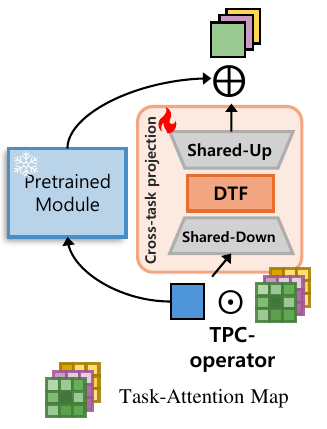}
    \caption{TADFormer}
    \label{fig:figure2b}
  \end{subfigure}
  \vskip -0.1in
  \caption{Comparison of MTLoRA~\cite{agiza2024mtlora} and TADFormer. This illustrates the part that extracts and processes task-adapted features. Both methods employ additional trainable down-up projection structures like LoRA~\cite{hu2021lora}, in parallel to the frozen pre-trained module. While MTLoRA introduces additional down-up projections for multiple tasks, TADFormer integrates Dynamic Task Filter (DTF), which uses dynamic convolution operations, with minimal additional parameters as well as the TPC-operator using task attention map.}
  \label{fig:figure2}
\end{figure}

Moreover, these approaches 
disregard cross-task relation that is crucial in boosting MTL performance~\cite{zhang2021survey,vandenhende2021multi}. 
Fig.~\ref{fig:figure2a} conceptually illustrates current PEFT module, which includes both task-shared and task-specific components~\cite{agiza2024mtlora}. 
This method processes the task-shared and task-specific modules in parallel and do not provide opportunities for task-specific features to interact via task-shared modules. 

\begin{figure}[t]
  \centering
  \includegraphics[width=0.9\linewidth]{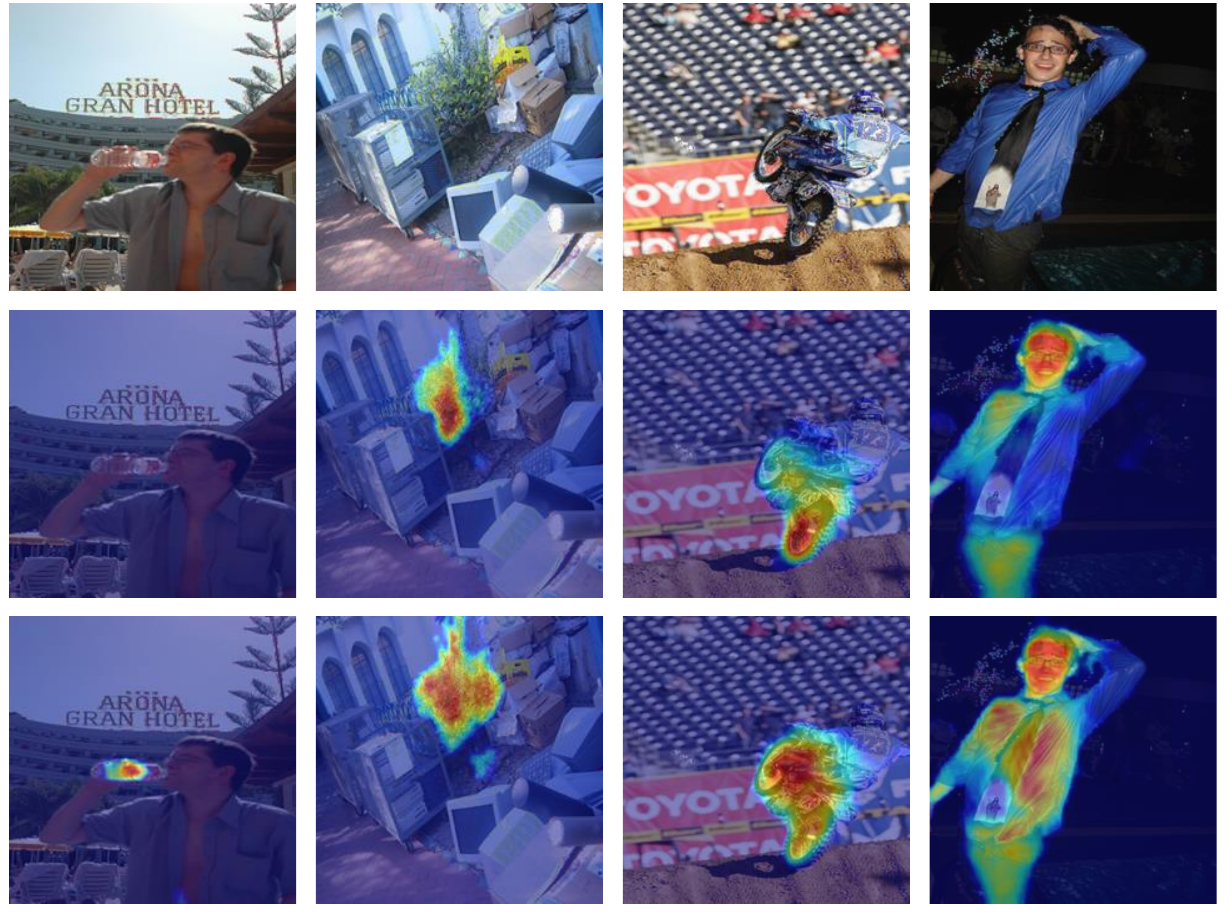}
  \vskip -0.1in
   \caption{Comparison of Grad-CAM~\cite{selvaraju2017grad} from MTLoRA~\cite{agiza2024mtlora} and TADFormer: (from top to bottom) 
   input images, MTLoRA, and TADFormer. This demonstrates that TADFormer is capable of extracting fine-grained features that capture the input contexts more precisely, thanks to DTF.
   }
   \label{fig:gradcam}
\end{figure}

To overcome these issues, we propose {\bf T}ask-{\bf A}daptive {\bf D}ynamic trans{\bf F}ormer, termed TADFormer, designed with two objectives; 1) to dynamically extract fine-grained task-specific features while considering interactions between task-specific features, and 2) to design a single module that efficiently extracts the task-specific features.
As shown in Fig.~\ref{fig:figure2b}, task-adapted features are generated with the help of task attention maps that represent task-specific attributes for each task.
They are then fed into a Task-Aware Module (TA-Module) augmented by Dynamic Task Filter (DTF). The DTF adaptively captures fine-grained features by leveraging the context information of input samples through dynamic convolution operations. Furthermore, the distinct task-adapted features are transmitted through the cross-task projection, which enables the model to consider cross-task relationships. 
Through the Grad-CAM visualization in Fig.~\ref{fig:gradcam}, we confirm that our model is capable of extracting fine-grained features more effectively when the input context is considered in the finetuning process.



 \textbf{Our contributions} can be summarized as follows: 


\begin{itemize}
\item \  
We propose TADFormer, a novel PEFT framework for multi-task scene understanding. TADFormer utilizes DTF to dynamically extract task-specific features based on the input context, allowing for more fine-grained task-specific feature learning. 
\item \ 
We propose a prompt-based task adaptation mechanism within the encoder. This approach significantly reduces the number of parameters while effectively learning task-specific features. 
\item \ 
TADFormer achieves higher accuracy while training fewer parameters compared to fully fine-tuning the entire model. TADFormer also demonstrates superior parameter efficiency and accuracy compared to recent PEFT methods as shown in Fig.~\ref{fig:figure1}.

\end{itemize}


\section{Related Work}
\label{sec:formatting}

\subsection{Multi-Task Learning (MTL)}
MTL aims to enable models to learn multiple related tasks simultaneously by sharing information unique to each task~\cite{zhang2021survey}. Effective MTL requires capturing useful information in three key areas: task-agnostic representations, task-specific representations, and cross-task interactions. To this end, various MTL approaches have been developed, generally categorized into encoder-focused and decoder-focused approaches~\cite{vandenhende2021multi}. The encoder-focused architectures primarily facilitate feature sharing within the encoder, using either hard or soft parameter sharing. In the hard parameter sharing strategy, multiple tasks share a common set of parameters in the encoder, while each task has an independent decoder to yield task-specific outputs~\cite{bekoulis2018adversarial,sener2018multitask,kendall2018multitask}. In contrast, the soft parameter sharing strategy allows each task to maintain its own set of encoder parameters and exchange cross-task information through modules integrated within the encoder~\cite{bhattacharjee2022mult,gao2019nddr}. The decoder-focused architectures~\cite{ye2022taskprompter,huang2024synergy,ye2023taskexpert,yang2024multi,kim2022sequential} extend cross-task interactions into the decoding stage. These models typically employ task-specific and cross-task modules within the decoder, enabling tasks to share a single encoder while enabling for more flexible information exchange in the decoders. 
\subsection{Parameter-Efficient Fine-Tuning (PEFT)}
PEFT aims to adapt large-scale pre-trained models~\cite{vit, swin, swinv2, segformer, detr, lee2022knn, lee2023cross, dpt, lee2022knn, lee2023cross, xie2022simmim, he2022masked, yi2022masked, choi2024emerging, choi2024improving, choi2025salience} to downstream tasks by tuning only a small subset of parameters, without re-training the entire model. 
Representative methods include Prompt tuning~\cite{lester2021prompt,li2021prefix}, Adapters~\cite{houlsby2019parameter,lei2023conditional}, and Low-Rank Adaptation (LoRA)~\cite{hu2021lora,zhong2024conv_lora}.
Prompt Tuning introduces trainable tokens to the input of the pre-trained model, facilitating task-specific adaptation without modifying the original weights. Adapters add compact, trainable modules to the model, with only these modules being fine-tuned to adapt the model to specific tasks. AdaptFormer~\cite{chen2022adaptformer} was the first adapter applied to computer vision and demonstrated the effectiveness of adapters in vision tasks by achieving competitive performance. In contrast, LoRA applies low-rank decomposition to the model’s weight matrices, enabling task-specific tuning while maintaining the original weight structure. 
LoRA is particularly efficient by introducing no additional parameters at inference time as the trainable matrices merge directly with the frozen weights. This allows for effective weight updates with minimal computational overhead.
Prompt tuning, Adapters, and LoRA have been successfully applied across various fields of NLP~\cite{houlsby2019parameter,hu2021lora,lester2021prompt} and computer vision~\cite{yin2023lowrank,chen2022adaptformer,jia2022visual}, demonstrating their versatility and efficiency. 
Additionally, LoRand~\cite{yin2023lowrank} introduces multi-branch low-rank adapters to improve performance on dense prediction tasks, and SPT~\cite{he2023sensitivity} further explores adaptive parameter allocation to task-specific important positions using tuning methods such as LoRA and Adapters.

\subsection{PEFT for MTL}
MTL experiences increased training complexity due to the proportional growth in decoder size with the number of tasks. To address this, recent research has been actively applying PEFT methods, originally designed for single-task scenarios, to the MTL model. Applying PEFT to multi-task models is challenging in that unique attributes of each task should be extracted through pre-trained models only with small amount of learnable parts. MTLoRA~\cite{agiza2024mtlora} and VMT-Adapter~\cite{xin2024vmt} have been proposed as PEFT approaches for MTL. MTLoRA utilizes LoRA modules consisting of task-agnostic and task-specific components, which effectively separate the parameter space during MTL fine-tuning. In contrast, VMT-Adapter enhances task interactions by sharing knowledge across tasks and preserves task-specific knowledge through independent knowledge extraction modules. 
However, these approaches struggle to account for the input context when extracting task-specific features, which limits their ability to capture fine-grained details. 


\section{Method}

\begin{figure*}
  \centering
  \begin{subfigure}{0.58\linewidth}
    \centering
    \includegraphics[width=\linewidth]{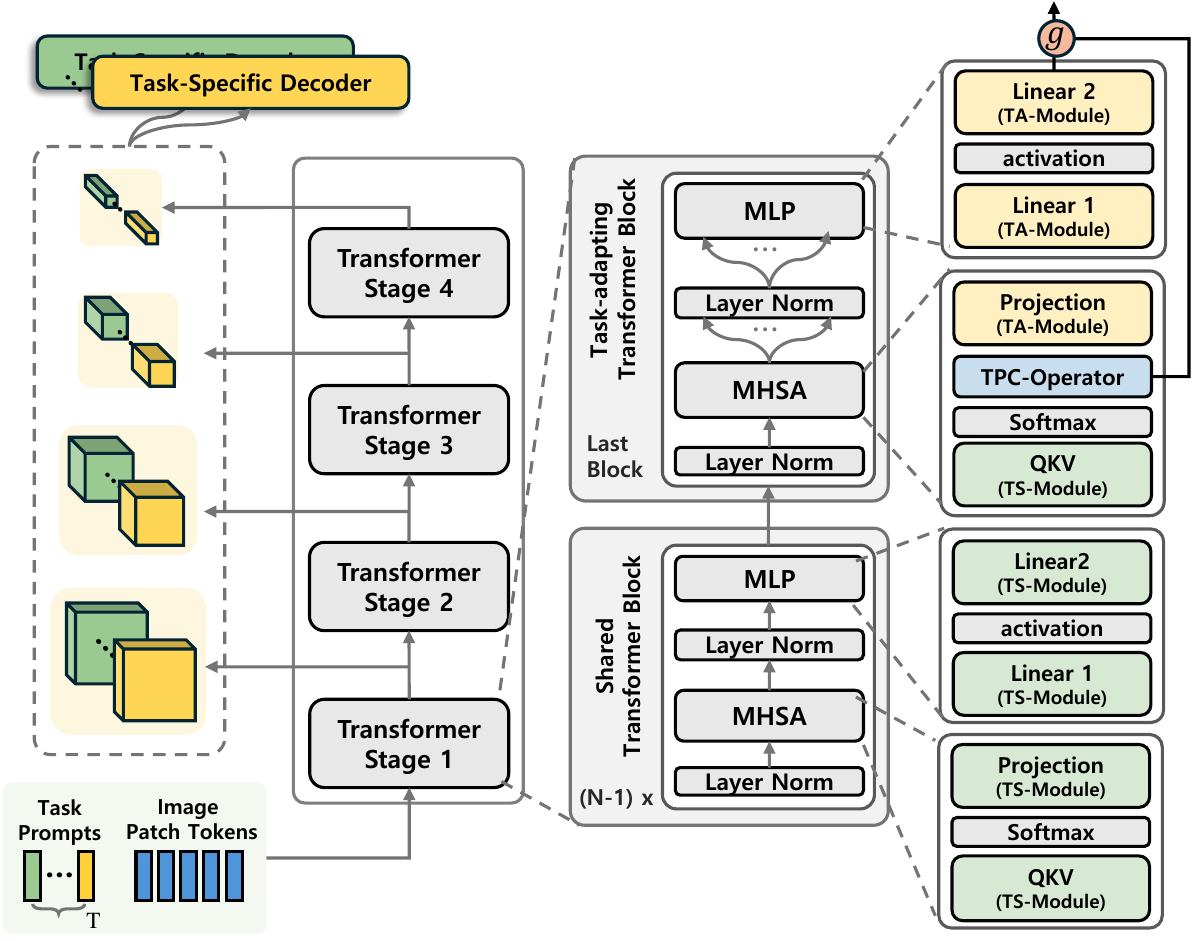}    
    \caption{Overall architecture of TADFormer}
    \label{fig:TADFormer_architecture}
  \end{subfigure} 
  \hfill
  \begin{subfigure}{0.38\linewidth}
    \centering
    \includegraphics[width=\linewidth]{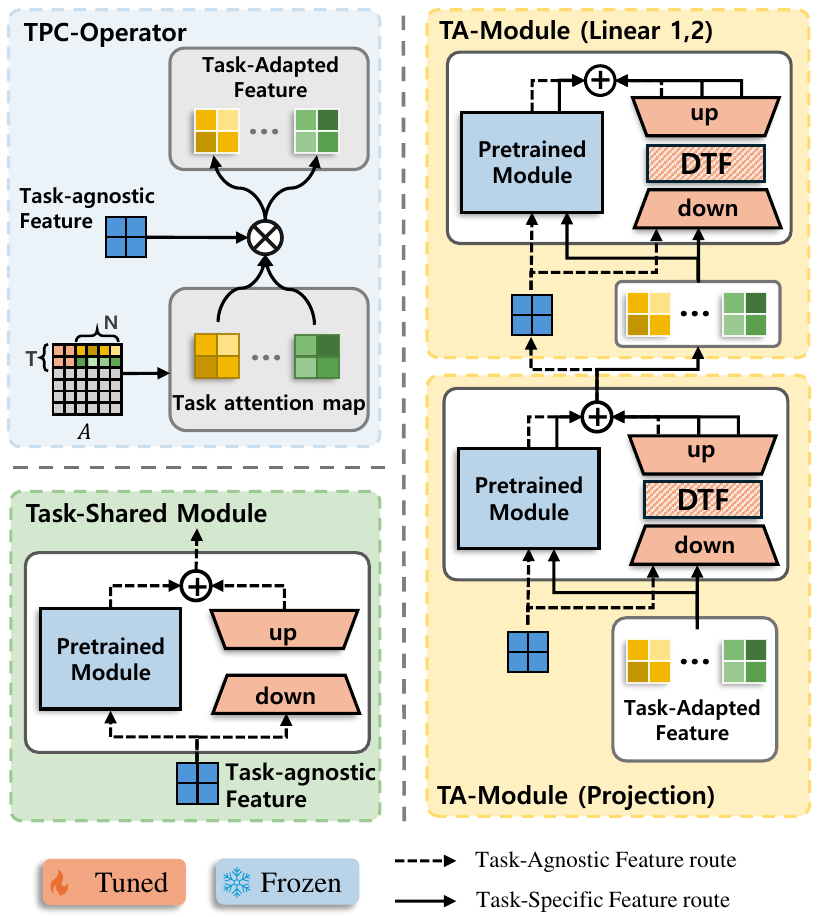} 
    \caption{TS-Module, TA-Module, and TPC Operator}
    \label{fig:Modules}
  \end{subfigure}

  \caption{Overview of the proposed TADFormer: (a) The encoder takes as inputs image patch tokens with task prompts prepended. Here, we adopt the VPT shallow approach \cite{jia2022visual}, inserting the task prompts only into the first Transformer stage, and use Swin Transformer~\cite{liu2021swin} as the encoder backbone. In all blocks except the last one of each Transformer stage, the task-agnostic features are extracted through the task-shared module (TS-module). The task-adapting Transformer block extracts fine-grained task-specific features through the task-prompt conditional (TPC) operator and task-aware module (TA-Module), (b) The TPC operator generates the task-adapted features with the help of the task attention map between task prompts and image patch tokens, and these features are then fed into the TA-module consisting of the dynamic task filter (DTF) as well as down-up projections for considering input contexts that are crucial to MTL.}
  \label{fig:TADFormer_overview}
\end{figure*}

\subsection{Overview} 
\label{Overview}
The proposed architecture, TADFormer, consists of a task-shared encoder with a series of Transformer layers and multiple task-specific decoders, as depicted in Fig.~\ref{fig:TADFormer_overview}. A set of task prompts $P=\{p^{ini}_1, ..., p^{ini}_T\}$ is prepended to image patch tokens $E=\{ e^{ini}_1, ...,e^{ini}_N \}$, forming $X=[P, E]$, and it is then fed into the task-shared Transformer encoder. $T$ and $N$ represent the number of tasks and patch tokens, respectively. Similar to current PEFT approaches for MTL \cite{agiza2024mtlora}, $N-1$ Transformer blocks of each stage are designed with task-shared modules (TS-Module) consisting of frozen pretrained parts and learnable LoRA modules \cite{hu2021lora}. In contrast, the last Transformer block, termed Task-Adapting Transformer block in Fig.~\ref{fig:TADFormer_architecture}, is defined with the proposed task-aware modules (TA-Module) that generate task-adaptive and input-dependent features through the task-prompt conditional (TPC) operator and dynamic task tilter (DTF).

\subsection{Task-Shared Module (TS-Module)}
To efficiently process task-agnostic feature in a parameter-efficient manner, a frozen pre-trained module $\Phi(\cdot)$ is decomposed into low-rank matrices with down-projection parameters $W_{down} \in \mathbb{R}^{C \times r}$ and up-projection parameters $W_{up} \in \mathbb{R}^{r \times C}$ \cite{hu2021lora}, where $r\ll C$. 
When an input feature $X_{in} \in \mathbb{R}^{(T+N) \times C}$ is given, a task-agnostic output feature $X_{out} \in \mathbb{R}^{(T+N) \times C}$ from the TS-module is defined as follows: 
\begin{equation}
    X_{out} = \Phi(X_{in}) + (X_{in}W_{down})W_{up}.
\label{Shared-Module}
\end{equation}

\noindent Similar to \cite{agiza2024mtlora} that uses LoRA \cite{hu2021lora} for efficient tuning in the MTL model, 
the TS-Module is applied to QKV layer, projection layer, and MLP block within the Transformer blocks, as shown in Fig.~\ref{fig:TADFormer_overview}. Note that while the TS-module is used in all Transformer blocks, the proposed TA-module and TPC operator with the aims of extracting distinct features for tasks are used only in the last Transformer block. This will be detailed in the following sections.

\subsection{Task-Prompt Conditional (TPC) Operator}
\label{Task-Prompt}

The TPC operator aims to effectively decouple task-specific features from the task-agnostic features, so that the DTF is capable of extracting representations that further reflects the characteristics of each task. 
Specifically, it enhances image features of high attention for each task by using task-prompts. 
Inspired by \cite{ye2022taskprompter,choi2025salience}, we propose to use a task attention map between task prompt and patch tokens derived directly from the  MHSA module of the Transformer backbone. 

\subsubsection{Task Attention Map}
Task prompts, which are tuned along with task-agnostic feature within the TS-module, are used to extract task-adapted features at the last Transformer block of each stage.
In the task-adapting Transformer block of Fig.~\ref{fig:TADFormer_architecture}, the QKV module generates the task-agnostic feature $f_{qkv} \in \mathbb{R}^{N \times C}$ and its associated attention map $A \in \mathbb{R}^{H \times (T+N) \times (T+N)}$, where $H$ is the number of heads in the multi-head self-attention (MHSA). 
The task attention map ${A}_{\text{TAM}}\in \mathbb{R}^{H \times T \times N}$ between the task prompts and patch tokens is simply obtained from $A$, such that ${A}_{\text{TAM}}=\left\{ a_1, ..., a_T \right\}$ and $a_i \in \mathbb{R}^{ H \times 1 \times N }$. Here, $a_i$ indicates how the task prompt $p_i$ relates to all patch tokens $E=\{e_1,...,e_N\}$. To match the dimension of the task-agnostic feature to the number of heads, we also reshape $f_{qkv}$ such that $\hat{F}=\mathrm{S}(F)\in \mathbb{R}^{H \times C/H \times N}$.

\subsubsection{Task-Adapted Feature}
The task-adapted feature $f_i\in \mathbb{R}^{N \times C}$ of the task $i$ for $i=1,...,T$ is computed as follows:
\begin{equation}
    f_i = f_{qkv} + \mathrm{S_{inv}} ({a_i\otimes \hat{f}_{qkv}}),
\label{task-adapted feature}
\end{equation}

\noindent where $\otimes$ is the Hadamard product, and it is repeatedly applied to $\hat{f}_{qkv}(\cdot,k,\cdot)$ for $k=1,...,C/H$. $\mathrm{S_{inv}}$ is another reshaping operator, which converts a matrix dimension from 
$\mathbb{R}^{H \times C/H \times N}$ to $\mathbb{R}^{N \times C}$.


\subsection{Task-Aware Module (TA-Module)}

The TA-Module is a main component of the TADformer, designed to effectively capture fine-grained task-specific features and perform task-cross interactions. As depicted in Fig.~\ref{fig:Modules}, the TA-module is structured with the Dynamic Task Filter (DTF) inserted between low-rank decomposed parameters used in the TS-Module. 
The TA-Module processes different inputs depending on its placement within the network. In the projection layer, it processes the task-adapted features, which are outputs of from the TCP-Operator. 
When applied to the linear layer, it processes the previous TA-module's output.


\begin{figure}
  \centering
  \includegraphics[width=0.6\linewidth]{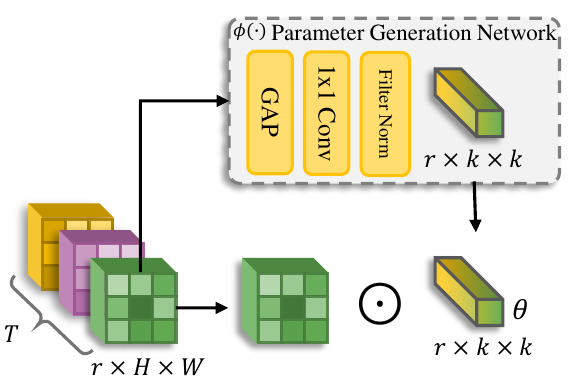}
  \vskip -0.1in
   \caption{DTF architecture: The down-projected features of channel dimension $r$ are used to generate channel-wise convolution parameters in the parameter generation network. GAP denotes global average pooling.}
   \label{fig:DTF}
\end{figure}

\subsubsection{Dynamic Task Filter (DTF)} 
\label{DTF}

Recent MTL studies~\cite{jia2016dynamic, yang2024multi} emphasize that the contextual information of input instances plays a crucial role in capturing the unique characteristics of each task.
Tuning only a subset of parameters to apply PEFT to multi-task models is likely to limit the model's ability to adapt to multiple tasks and capture task-specific features. 
To address this challenge, we propose to use the DTF that considers both input and task context within the TA-module, enabling more effective extraction of fine-grained task-specific features.
The key objective of DTF is to generate task-customized convolutional parameters $\theta = \left\{ \theta_1, ..., \theta_T \right\}$ that reflects the context of the input task-adapted features $f = \left\{ f_1, ..., f_T \right\} $ while maintaining trainable parameter-efficiency.


A naive approach for generating task-customized parameters is to directly input the feature $f_i$ into the parameter generation network $\phi(\cdot)$, such that $\theta_i = \phi(f_i)$. However, it requires processing the entire feature to generate $\theta$, failing to effectively balance trainable parameter efficiency and performance. Instead, we adopt more efficient strategy, in which $\phi(\cdot)$ generates channel-wise convolution parameters $\theta$. 
Additionally, we apply Global Average Pooling (GAP) to the input features, resulting in a more lightweight network, as depicted in Fig.~\ref{fig:DTF}. This approach reduces the number of parameters to $r \times r \times k^2$, where $k$ is the kernel size. The final output $\Tilde{F}$ of the TA-module using the DTF is as follows:
\begin{equation}
    \theta_i = \phi(f_i W_{down})
\label{DTF-parameter generation}
\vspace{+0.6em} 
\end{equation}
\begin{equation}
    \Tilde{F}_i = \Phi(f_i) + (\theta_i \odot (f_i W_{down}))W_{up}
\label{DTF-Final output}
\end{equation}
Here, $\odot$ is a channel-wise convolution operations.
Additionally, since DTF dynamically adapts to both input and task contexts, training may become unstable, so we employ FilterNorm \cite{zhou2021decoupled} to the parameter generation network for securing training stability.

\subsection{Stage-wise Gating and Skip Connection}
The task-wise feature adaptation is conducted through the TPC operator and TS-modules within the block. 
To further improve usability of the task-adapted feature $f_i$ ($i=1,...,T$) in the task-specific decoders, we introduce the skip connection that adds it to the last Transformer block output $\hat{f_i}$, which is then fed to the task-specific decoders, as shown in Fig.~\ref{fig:TADFormer_architecture}.
This skip connection allows the task-adapted feature $f_i$, generated through the task prompts in the TPC operator, to be directly passed to the task-specific decoder, allowing the task prompt to better capture attention related to its corresponding task. Additionally, we incorporate a stage-wise gating with a parameter $g$ in the skip connection.
A final output of each Transformer stage $F_i$ is as follows:
\begin{equation}
    F_i = \sigma(g) \cdot  f_i + (1-\sigma(g)) \cdot  \hat{f_i},
\label{gating and layering}
\end{equation}
\noindent where $\sigma$ indicate the Sigmoid activation function. A trainable gating parameter $g$ is initialized to zero and regulated by the Sigmoid function.


\subsection{Selection of Fine-tuning Modules}

\label{Lightweight}

\begin{figure}
  \centering
  \includegraphics[width=0.7\linewidth]{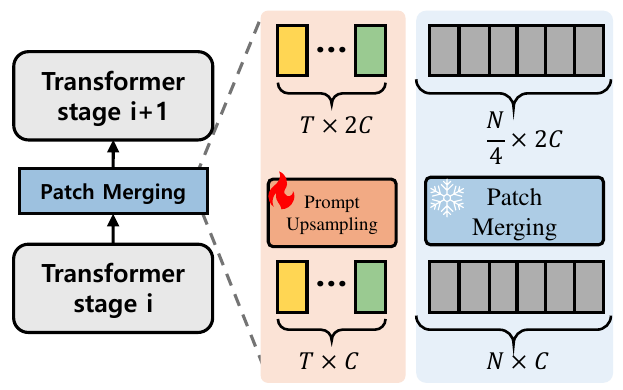}
  \vskip -0.1in
   \caption{Patch merging module of TADFormer: Leveraging the benefits of task-prompt tuning, we finetune only the prompt upsampling operation that doubles the channel size of the task prompts, while freezing the image patch merging module.}
   \label{fig:prompt upsampling}
\end{figure}

In the PEFT domain, several studies have been conducted to select which modules are tuned to achieve the best trade-off between downstream task accuracy and efficiency~\cite{agiza2024mtlora,gao2023llama}. 
For example, \cite{agiza2024mtlora} demonstrated that tuning non-attention modules, such as layer normalization, patch merging layers, and patch embedding layer achieves an effective trade-off in utilizing low-rank adaptation in the hierarchical Transformer baseline, like Swin Transformer~\cite{liu2021swin}. In this regard, we explore an optimal selection of tuning modules in our TADFormer that leverages both low-rank adaptive modules and task-prompt tuning.

Since our approach is based on the shallow-VPT framework \cite{jia2022visual} where task prompts are not newly added in the middle of layers, an prompt-upsampling module that doubles the channel size of task prompts is necessary when passing them to the next Transformer stage.
Considering the prompt is upsampled in the same position as the patch merging module, we freeze the patch merging module and tune only the prompt upsampling module, as shown in Fig.~\ref{fig:prompt upsampling}.
By tuning only the prompt upsampling module with  $2C \times C$ trainable parameters, we achieved competitive performance without additionally tuning the patch merging module with $4C \times 2C$ trainable parameters. 
Additionally, we tune the layer normalization and positional embedding layers, similar to \cite{agiza2024mtlora}.





\begin{table*}[t]
  \centering
  \setlength{\tabcolsep}{4.5pt}
\renewcommand{\arraystretch}{0.85}
  \vspace{-5pt}
  \caption{Performance comparison 
  using Swin-T pretrained on ImageNet-1k as the backbone. Here, $\Delta m$ denotes the relative improvement compared to single-task full fine-tuning.  $\uparrow$ and $\downarrow$ indicate whether higher or lower values are preferable, respectively. Bold values indicate that TADFormer yields approximately 1.2-1.7\% higher $\Delta m$ with fewer parameters than MTLoRA.} 
\small
  \begin{tabular}{c | c c c c | c | c }
    \toprule
    \multirow{2}{*}{\textbf{Method}} & \textbf{SemSeg} & \textbf{Human Parts} & \textbf{Saliency} & \textbf{Normals} & 
      \multirow{2}{*}{$\Delta m (\%) $} & \textbf{Trainable}  \\
      & ($mIoU \uparrow$) & ($mIoU \uparrow$) & ($mIoU \uparrow$) & ($rmse \downarrow$) & &
      \textbf{Parameters} (M) \\
    \midrule
    Single Task & 67.21 & 61.93 & 62.35 & 17.97 & 0 & 112.62 \\
    MTL - Tuning Decoders Only & 65.09 & 53.48 & 57.46 & 20.69 & -9.95  & 1.94 \\
    MTL - Full Fine Tuning  &  67.56 & 60.24 & 65.21 & 16.64 & +2.23 &  30.06   \\
    \midrule
    Adapter \cite{he2021towards} &  69.21 &  57.38 & 61.28 & 18.83  & -2.71 & 11.24 \\
    Bitfit \cite{zaken2021bitfit} & 68.57 & 55.99 & 60.64 & 19.42 & -4.60 & 2.85   \\
    VPT-shallow \cite{jia2022visual}  & 62.96 &  52.27 & 58.31 & 20.90 & -11.18  & 2.57  \\
    VPT-deep \cite{jia2022visual}  & 64.35 & 52.54 & 58.15 & 21.07 & -10.85 & 3.43  \\
    Compactor \cite{mahabadi2021compacter} & 68.08 & 56.41 & 60.08 & 19.22 & -4.55 & 2.78  \\
    Compactor++ \cite{mahabadi2021compacter}  & 67.26 & 55.69 & 59.47 & 19.54 & -5.84 & 2.66 \\
    LoRA \cite{hu2021lora} & 70.12 & 57.73 & 61.90 & 18.96 & -2.17 & 2.87   \\
    VL-Adapter \cite{sung2022vladapter} & 70.21 & 59.15 & 62.29 & 19.26 & -1.83  & 4.74 \\
    HyperFormer \cite{mahabadi2021parameter}  & 71.43 & 60.73 & 65.54 &  17.77 & +2.64  & 72.77  \\
    Polyhistor \cite{liu2022polyhistor} & 70.87  & 59.54  & 65.47 &  17.47 & +2.34  & 8.96 \\
    MTLoRA \cite{agiza2024mtlora} ($r=16$) & 68.19 & 58.99 & 64.48 & 17.03  & +1.35 & 4.95\\
    MTLoRA \cite{agiza2024mtlora} ($r=32$) & 67.74 & 59.46 & 64.90 & 16.59 & +2.16 & 6.08\\
    MTLoRA \cite{agiza2024mtlora} ($r=64$) & 67.9 & 59.84 & 65.40 & 16.60 & +2.55 & 8.34\\
    \midrule
    TADFormer ($r=16$) & 69.79 & 59.27 & 65.04 & 16.91  & \textbf{+2.44} & \textbf{3.56} \\
    TADFormer ($r=32$) & 70.2 & 60 & 65.71 & 16.57 & \textbf{+3.63} & \textbf{4.78} \\
    TADFormer ($r=64$) & 70.82 & 60.45 & 65.88 & 16.48 & \textbf{+4.24}  & \textbf{7.38} \\
    \midrule
    \bottomrule
  \end{tabular}   
  
  \label{tab:Quantitive}
\end{table*}

\section{Experiments}
\subsection{Implementation Details}

\noindent \textbf{Dataset and Evaluation metrics}. To evaluate our proposed method, we conducted experiments on the PASCAL-Context dataset, following prior works on PEFT adaptation for MTL \cite{agiza2024mtlora, liu2022polyhistor}.
This dataset includes four types of annotations for dense prediction tasks: semantic segmentation with 21 classes, human part segmentation with 7 classes, surface normals estimation, and saliency detection. It consists 4,998 images for training and 5,105 images for testing, respectively. For evaluation, we used the mean intersection-over-union (mIoU) for semantic segmentation, human part segmentation, and saliency detection, and the root mean square error (rmse) for surface normals estimation. Additionally, to assess overall performance across tasks, we measured relative improvement over the single-task fine-tuning baseline as follows: 
\begin{equation}
    \Delta m = \frac{1}{T} \sum_{i=1}^{T} (-1)^l_i(M_{m,i} - M_{s,i})/M_{s, i},
\label{delta_M}
\end{equation}
where $T$ denotes the number of tasks, $M_{s,i}$ is the single task baseline and $M_{m,i}$ is the performance of the MTL models on the $i$-th task. Here, $l_i$ is set to 1 if lower value indicate better performance, otherwise 0.

\noindent \textbf{Implementation}. 
We used the Swin-Transformer pre-trained on the ImageNet dataset~\cite{deng2009imagenet} as the encoder, and HRNet~\cite{sun2019deep} as the decoder for a fair comparison with existing methods.
The decomposed matrices used in TS-Module and TA-Module are configured with ranks of 16, 32, and 64. Also, TS-Module and TA-Module use the same rank size. 
All experiments were performed in the Pytorch with the same experimental setup as~\cite{agiza2024mtlora}. The experiments in this work are all run on a single NVIDIA V100 GPU environment.

\noindent \textbf{Training}. For training the MTL model, we calculated the total loss through weighted sum using weight values corresponding to each task loss as in \eqref{Loss}. We used the task weights given in~\cite{vandenhende2020mti}.
\begin{equation}
    L_{MTL} = \sum_{i}^{T}w_i \times L_i
\label{Loss}
\end{equation}
where $w_i$ and $L_i$ are the weight and loss of task $i$.
As the loss for each task, standard per-pixel cross-entropy is used for semantic segmentation and human parts segmentation. L1 loss is used for surface normals estimation, and balanced cross-entropy loss is used for saliency detection.

\subsection{Baselines}
We compared the downstream task accuracy and number of trainable parameters with prior PEFT methods.
\textbf{Single-task full fine-tuning } is a full fine-tuning model that uses an individual pretrained model for each task.
\textbf{Adapter~\cite{he2021towards}} applies task-specific bottlenect modules to each Transformer layer.
\textbf{Bitfit}~\cite{zaken2021bitfit} tunes only the biases, patch merging layers and patch projection layers.
\textbf{VPT}~\cite{jia2022visual} tunes the model by prepending a trainable embedding, either to the input (VPT-shallow) or all transformer layers (VPT-deep).
\textbf{Compacter}~\cite{mahabadi2021compacter} decomposes the fast matrix into two low-rank vectors, while \textbf{Compacter++} places the modules exclusively after the MLP layers.
\textbf{LoRA}~\cite{hu2021lora} applies low-rank decomposition to the attention layers, using a rank of 4 and an adapter output scale of 4.
\textbf{VL-Adapter}~\cite{sung2022vladapter} simply adds a single adapter to the transformer and shares it across tasks.
\textbf{Hyperformer}~\cite{mahabadi2021parameter} uses a hyper-network that generates adapter weights for different tasks based on task embeddings.
\textbf{Polyhistor}~\cite{liu2022polyhistor} introduces low-rank hypernetworks and custom kernels to scale fine-tuning parameters across Transformer blocks.
\textbf{MTLoRA~\cite{agiza2024mtlora}} utilizes a dual-module approach that uses both task-agnostic LoRA modules and task-specific LoRA modules. Task-agnostic LoRA modules capture task-generic features, and task-specific LoRA modules tailor task-specific features from this shared representation.

\subsection{Quantitative Analysis}

\begin{table*}
  \centering
  \setlength{\tabcolsep}{8pt} 
  \renewcommand{\arraystretch}{1.2} 
  \caption{\textbf{Ablation study}: `TADFormer Baseline' refers to the case where all Transformer blocks including the last one utilize only the TS-modules in Fig.~\ref{fig:TADFormer_architecture}. `+ TPC-operator + TP' indicates the performance when the TPC-operator is used with the task prompts (TP) but the DTF is not used. Inversely, `+ DTF' refers to the case that the DTF of the TA-module receives the same features for all tasks, without the TPC-operator and the task prompts. 
  `TP + DTF’ includes the DTF and tuning the task prompts, but does not apply TCP-operator using task attention map (TAM) computed from the task prompts.
  }
  \small
  \begin{tabular}{l | c c c c | c | c }
    \toprule
    \multirow{2}{*}{\textbf{Method}} & \textbf{SemSeg} & \textbf{Human Parts} & \textbf{Saliency} & \textbf{Normals} & 
      \multirow{2}{*}{$\Delta m (\%) $} & \textbf{Trainable}  \\
      & ($mIoU \uparrow$) & ($mIoU \uparrow$) & ($mIoU \uparrow$) & ($rmse \downarrow$) & &
      \textbf{Param.} (M) \\
    \midrule
    Single Task & 67.21 & 61.93 & 62.35 & 17.97 & 0 & 112.62 \\
    \midrule
   TADFormer Baseline & 68.16 & 59.14 & 64.72 & 17.16 & +1.3 & 4.26  \\  
    + TPC-operator + TP (A) & 68.9 \textbf{($\uparrow$0.74)}& 59.1 \textbf{($\downarrow$0.04)} & 64.82 \textbf{($\uparrow$0.1)} & 17.03 \textbf{($\downarrow$0.13)} & +1.79 \textbf{($\uparrow$0.49)} & 4.66 \\
    + DTF (B) & 70.17 \textbf{($\uparrow$2.01)} & 59.82 \textbf{($\uparrow$0.68)} & 65.24 \textbf{($\uparrow$0.52)} & 16.86 \textbf{($\downarrow$0.3)} & +2.95 \textbf{($\uparrow$1.65)} & 4.40\\
    + TP + DTF & 70.29 \textbf{($\uparrow$2.13)} & 59.8 \textbf{($\uparrow$0.66)} & 65.5 \textbf{($\uparrow$0.78)} & 16.75 \textbf{($\downarrow$0.41)} & +3.24 \textbf{($\uparrow$1.94)} & 4.78\\
    TADFormer (A+B) & 70.2 \textbf{($\uparrow$2.04)} & 60 \textbf{($\uparrow$0.86) }& 65.71 \textbf{($\uparrow$0.99)}& 16.57 \textbf{($\downarrow$0.59)} & +3.63 \textbf{($\uparrow$2.33)} & 4.78 \\
    \bottomrule
  \end{tabular}
  \label{tab:ablation}
\end{table*}

\begin{figure}
  \centering
  \includegraphics[width=1\linewidth]{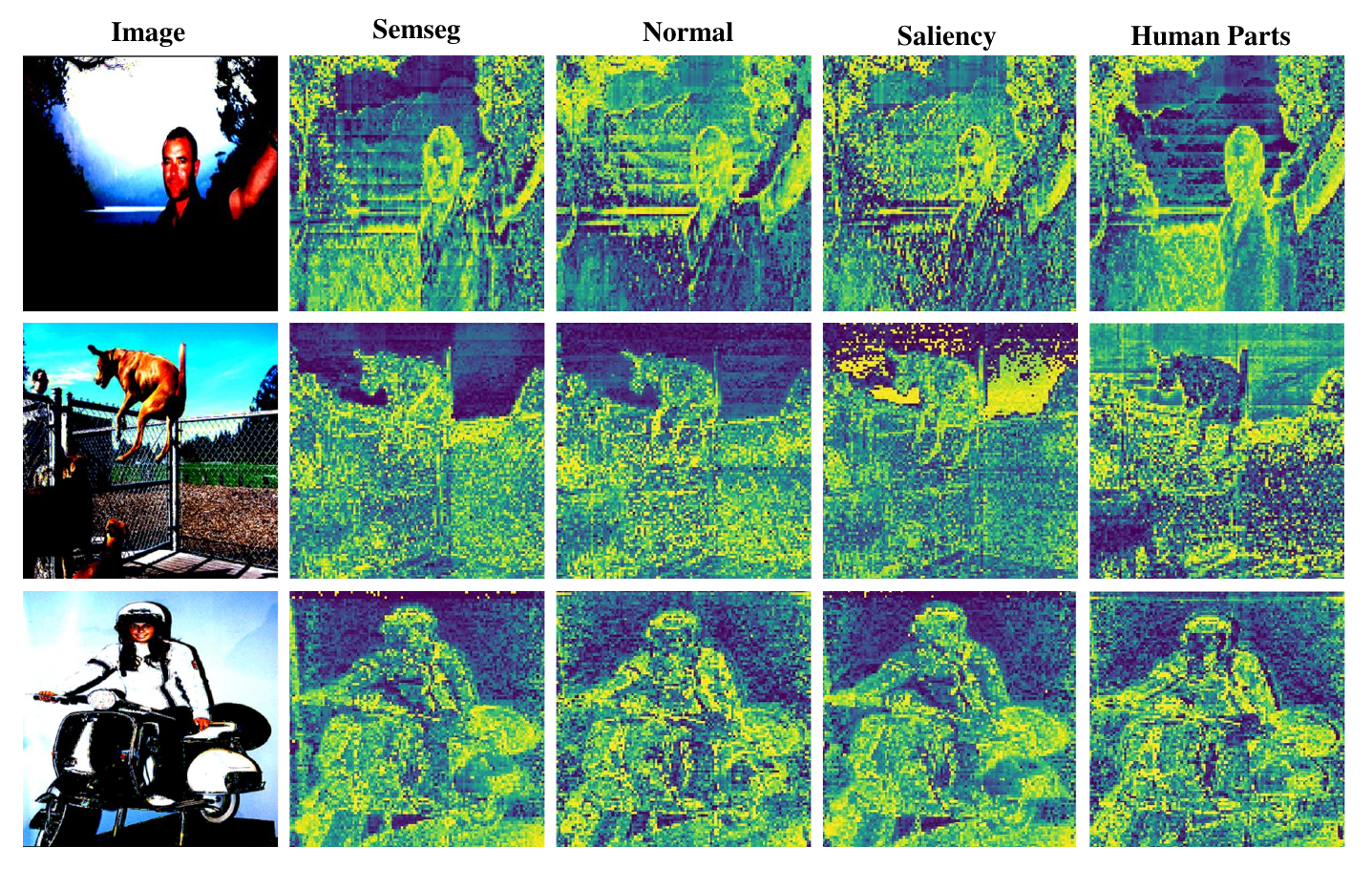} 
  \caption{Visualization of the task attentnion map (TAM) used by the TPC-operator. Notice that the task-prompts corresponding to each task are focused on different parts of the image.} 
  \label{fig:TAM}
\end{figure}

Table~\ref{tab:Quantitive} shows the MTL performance of all other baseline PEFT methods, including trainable parameters of each methods. For fair comparison, all methods used the Swin-T model~\cite{liu2021swin} pretrained on ImageNet-1k as the backbone.
When adapting PEFT to MTL downstream tasks, it is essential to consider 1) the overall multi-task downstream accuracy with 2) the number of trainable parameters relative to the full model parameters. In Table~\ref{tab:Quantitive}, we highlight the results of TADFormer in bold that outperform that the state-of-the-arts \cite{agiza2024mtlora}, while utilizing fewer parameters. These results, reinforced by the structural differences shown in Fig.~\ref{fig:figure2}, demonstrate that TADFormer is trained in a more parameter-efficient manner, making it highly suitable for MTL.

\subsection{Ablation Study}
The impact of the main modules in TADFormer is reported in the ablation study of Table~\ref{tab:ablation}. \textbf{1)} `\textbf{TADFormer Baseline}' refers to the configuration that utilizes only the TS-modules for all Transformer blocks including the last one, without task-prompt tuning with task prompts prepended to the image patch tokens.
\textbf{2)} `\textbf{+ TPC-operator + TP}' represents the configuration when the `TADFormer Baseline' includes task-prompt tuning (TP) and task-adapting via TCP-operator without using the DTF.
\textbf{3)} `\textbf{+ DTF}' is the model where the DTF is added without the TPC-operator. This model does not include the task prompts as in the TADFormer baseline.
\textbf{4)} `\textbf{+ TP + DTF}' the DTF and tuning the task prompts (TP), but does not apply TCP-operator using task attention map (TAM) computed from the task prompts. This demonstrate the effectiveness of task attention map (TAM) in TPC-operator to tailor task-adapted features from task-agnostic feature.
\textbf{5)} \textbf{TADFormer} means our proposed method including all components.
The low-rank $r$ is fixed to 32 in all the experiments utilized in Ablation. 



\noindent \textbf{Effectiveness of the TPC-operator and DTF}. 
Comparing the performance of `\textbf{TADFormer Baseline}' and `\textbf{+ DTF}' in Table~\ref{tab:ablation}, we can see that the DTF can increase $\Delta m$ by 1.65. Also, note that `\textbf{+ DTF}' already yields a high performance improvement despite not considering task-context since it does not separate task-adapted features through the TPC. This confirms that the process of extracting input-context sensitive representations via the DTF also is a core component in the MTL performance. 
Comparing the TADFormer Baseline and `\textbf{+ TPC operator + TP}' in Table~\ref{tab:ablation}, we can see an increase in $\Delta m$ of 0.49. This confirms that the task-adapted feature tailoring process via the task-attention map is significant. 
The highest $\Delta m$ performance is achieved by using both TPC-operator with the task prompts and DTF in `\textbf{TADFormer}'. When the two main components are used together, the TPC-operator passes the task contexts to the DTF. From this, DTF is capable of extracting fine-grained task specific features that take into account both input and task contexts.

\noindent \textbf{Effectiveness of Task Attention Map (TAM)}
Table~\ref{tab:ablation} also shows the effectiveness of generating task-adapted features by using the TCP-operator based on the task attention map (TAM). `\textbf{TADFormer}' and `\textbf{+ TP + DTF}' have the same number of trainable parameters, but 0.39 of the $\Delta m$ is reduced when eliminating the TCP-operator based on the TAM. This confirms the significance of incorporating task-context features into the model via the TAM. Also, the TAM visualization in Fig.~\ref{fig:TAM} shows that the task prompts are able to capture different regions of attentions for each task.




\begin{figure}
  \centering
  \includegraphics[width=0.9\linewidth]{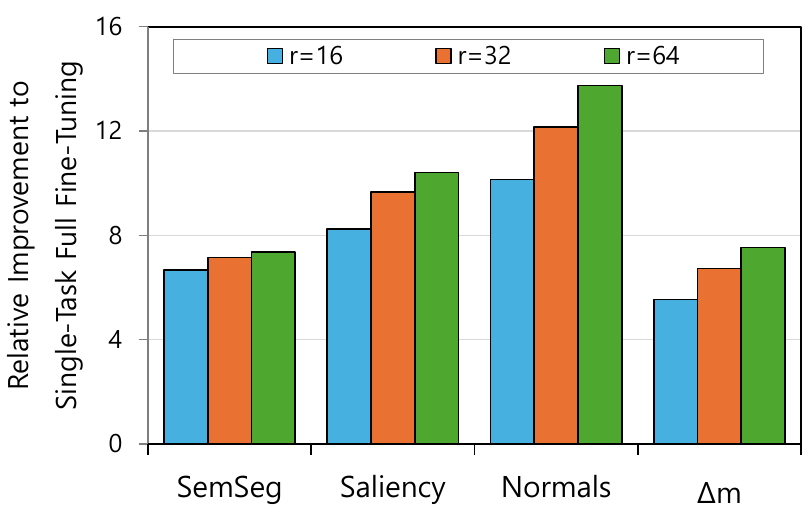}
  \vskip -0.1in
   \caption{ Performance of TADFormer when employing the Swin-Basedel pre-trained on ImageNet-22k. }
   \label{fig:Base22k}
\end{figure}

\noindent \textbf{Analysis on other Backbone and  Pre-training Dataset}.
To verify the applicability of the proposed method to other large backbone models and pre-training datasets, we applied TADFormer to Swin-Base pre-trained on the ImageNet-22k dataset. 
As shown in Fig.~\ref{fig:Base22k}, our method demonstrates greater performance improvements when using a larger feature backbone and more pretraining data. This suggests the potential for further performance gains by applying our method to larger model architectures and utilizing additional pretraining data.
More experiments are provided in the supplementary material.

\noindent \textbf{Extension to Adapter Baseline}.
While the proposed method is built upon the LoRA baseline~\cite{hu2021lora}, it can also be extended into adapter baselines~\cite{chen2022adaptformer, xin2024vmt} that are widely adopted recently in PEFT approaches due to low training complexity. These results will be provided in the supplementary material.



\section{Conclusion}
We have presented a novel PEFT approach that efficiently trains the MTL model for dense scene understanding tasks. To capture fine-grained, task-specific features that are crucial to improving the MTL model, we introduced TPC-operator based on parameter-efficient task prompting and TA-module using the DTF that is capable of reflecting task information conditioned on input contexts. Experiments on four representative tasks, including semantic segmentation, human part segmentation, surface normals estimation, and saliency detection, demonstrate outstanding performance, while keeping training complexity low. Future research includes the extension of this work by leveraging structural knowledge that exists among visual tasks~\cite{zamir2018taskonomy}.

\clearpage
\setcounter{page}{1}
\maketitlesupplementary

In the supplementary material, we provide more comprehensive results as follows.
\begin{itemize}
    \item More results using other backbone, pretraining dataset, and decoders
    \item Analysis on computational efficiency
    \item TADFormer on adapter-based methods
\end{itemize}

\section{More Results}

\subsection{TADFormer using other Backbone and Pre-training Dataset}
\label{sec:22k_Base}

\begin{figure}
  \centering
  \begin{subfigure}[b]{\linewidth}
    \centering
    \includegraphics[width=0.7\linewidth]{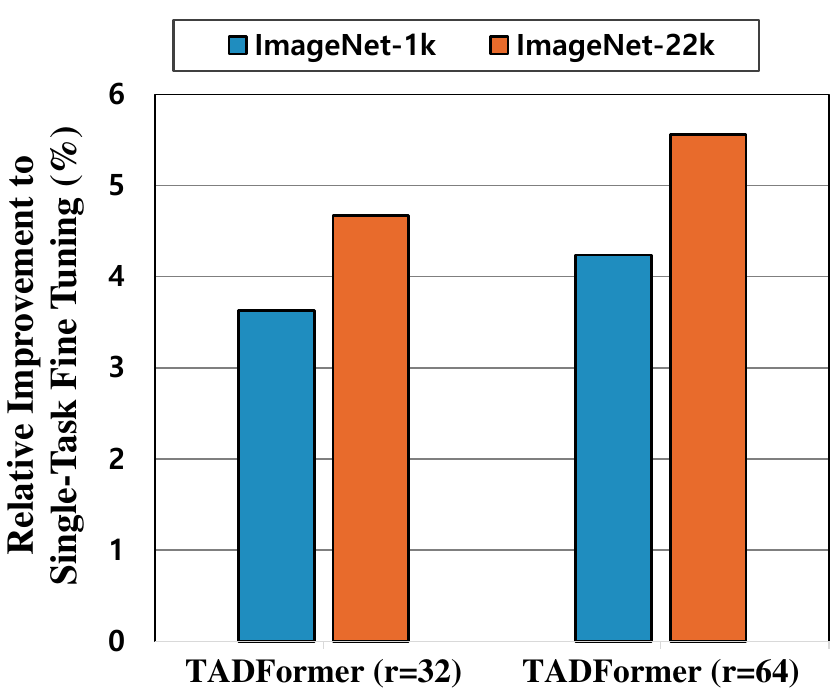}
    \caption{ImageNet-1k vs ImageNet-22k}
    \label{fig:22k}
  \end{subfigure}
  \vskip 1em
  \begin{subfigure}[b]{\linewidth}
    \centering
    \includegraphics[width=0.7\linewidth]{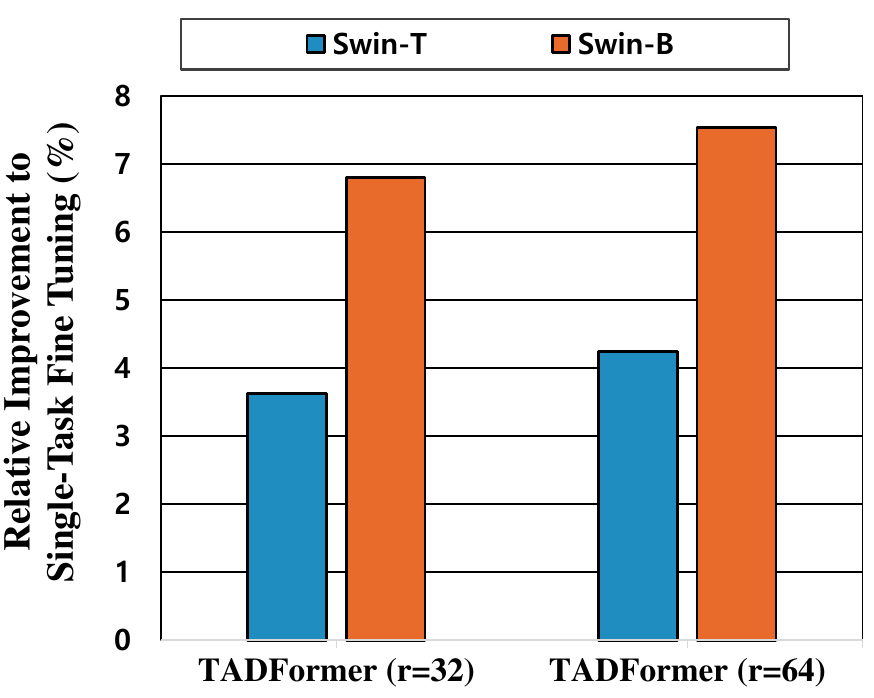}
    \caption{Swin-T vs Swin-B}
    \label{fig:base}
  \end{subfigure}
  \vskip -0.1in
  \caption{(a) illustrates the performance difference when using Swin-T pretrained on ImageNet-1k and ImageNet-22k as the backbone of TADFormer. (b) illustrates the performance variation of TADFormer when employing Swin-T and Swin-B, both pretrained on ImageNet-1k.}
  \label{fig:22k_base}
\end{figure}

Extending the experiment in Fig.~\ref{fig:Base22k}, we evaluated the performance of TADFormer using larger pretraining dataset and backbone. As shown in Fig.~\ref{fig:22k_base}, using the larger pretraining dataset improves performance significantly. 
Similarly, using the larger backbone, such as Swin-B, also resulted in improved performance.
It is worth of noting that the higher relative improvement is achieved when the larger back (Swin-B) or training dataset (ImageNet-22k) are used.
These results support the effectiveness of our approach, considering that the performance of single-task fine-tuning, which was used to compute the relative performance improvement, also improves.
This demonstrates that TADFormer is the scalable method that can effectively adapt to backbones and pretraining datasets of varying sizes.

\subsection{TADFormer using Other Decoders}
\label{sec:22k_Base}
We further evaluated the performance of TADFormer in combination with other decoders. We also compared it with the performance of MTLoRA when using the same decoder. Various decoders commonly used in dense prediction tasks were adopted, including HRNet~\cite{sun2019deep}, SegFormer~\cite{xie2021segformer}, and Atrous Spatial Pyramid Pooling (ASPP)~\cite{chen2018encoder}. The Swin-T pretrained on ImageNet-22k was used as the encoder.  
As shown in Table~\ref{tab:decoder}, TADFormer demontrates superior multi-task learning performance with fewer trainable parameters compared to MTLoRA across all decoder configurations, confirming that our method is flexible and can be integrated with various decoder architectures.
Additionally, ASPP shows the best performance as the decoder with the largest number of trainable parameters, indicating that the choice of decoder enables for an effective trade-off between the performance and the number of trainable parameters.

\begin{figure}
  \centering
  \begin{subfigure}[b]{\linewidth}
    \centering
    \includegraphics[width=0.7\linewidth]{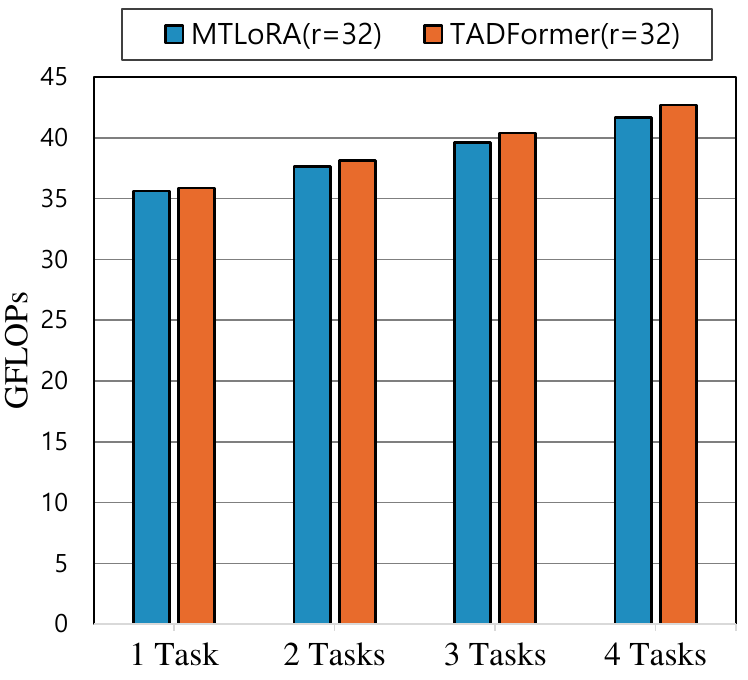}
    \caption{Comparison of GFLOPs between MTLoRA ($r=32$) and TADFormer ($r=32$) with respect to the number of tasks.}
    \label{fig:Supple_GFlops}
  \end{subfigure}
  \vskip 1em
  \begin{subfigure}[b]{\linewidth}
    \centering
    \includegraphics[width=0.7\linewidth]{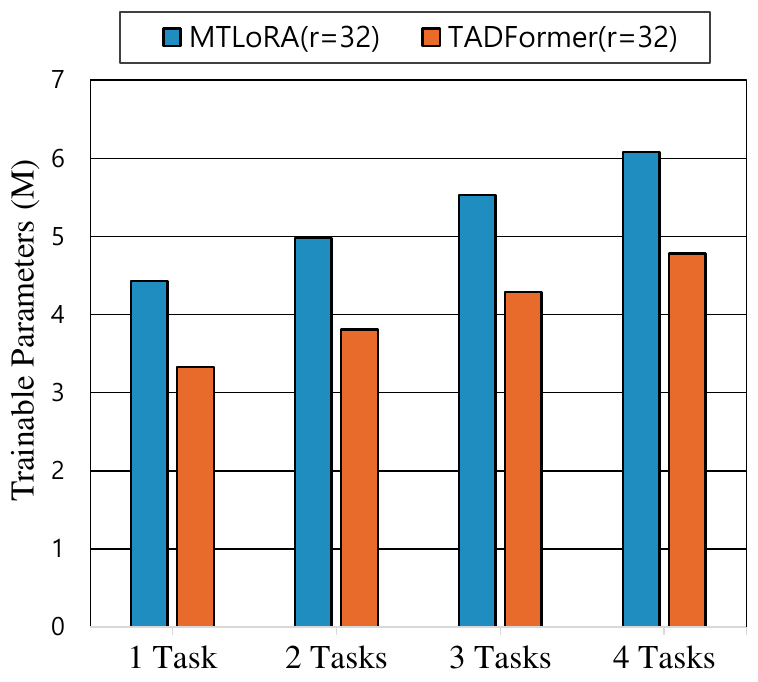}
    \caption{Comparison of the number of trainable parameters between MTLoRA ($r=32$) and TADFormer ($r=32$) with respect to the number of tasks.}
    \label{fig:Supple_params}
  \end{subfigure}
  \vskip -0.1in
  \caption{\textbf{Efficiency of TADFormer with different number of tasks}: The experiments were run at the rank $r=32$.}
  \label{fig:GFLOPs}
\end{figure}

\section{Analysis on Computational Efficiency}
\label{sec:Flops}

Fig.~\ref{fig:GFLOPs} shows the analysis on the computational efficiency in terms of GFLOPs and the number of trainable parameters with respect to the number of tasks. 
The GFLOPs for both MTLoRA and TADFormer are directly measeured in this experiment.
Compared to MTLoRA~\cite{agiza2024mtlora}, TADFormer requires slightly more GFLOPs but significantly fewer trainable parameters. This is because the TCP-operator in TADFormer does not require additional parameters, but instead needs additional computation for extracting task-adapted features, and DTF requires additional operations for parameter generation. 
While TADFormer has a marginal increase in GFLOPs, the substantial reduction in trainable parameters demonstrates its scalability and suitability for efficient multi-task learning scenarios, especially as the number of tasks increases. This trade-off indicates the efficiency of TADFormer in balancing training complexity and parameter optimization.


\section{TADFormer on Adapter-Based Methods}
\label{sec:adapter}

\begin{figure}
  \centering
  \includegraphics[width=0.8\linewidth]{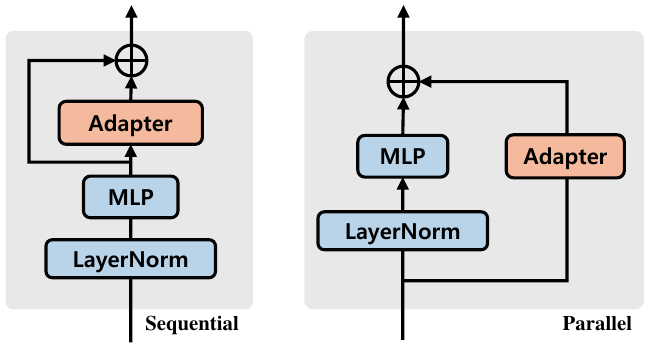} 
  \caption{Adapter configuration: Both sequential and parallel configurations are possible in adapter-based PEFT framework~\cite{chen2022adaptformer}.} 
  \label{fig:Adapter_config}
\end{figure}

\noindent\textbf{Experimental Setup}.
To analyze the extensibility of TADFormer into other PEFT methods, we experimented TADFormer with adapter-based methods~\cite{xin2024vmt,chen2022adaptformer}. The comparative analysis is shown in Table~\ref{tab:adapter}. 
The first column (Index) identifies the structures of the modules used in the experiments.
The AdaptFormer~\cite{chen2022adaptformer} is the adapter-based PEFT method for single task (\textbf{A1}$-$\textbf{A4}). The VMT-Adapter~\cite{xin2024vmt} extends the Adapter for MTL in a way that employs the task-shared adapter, similar to AdaptFormer~\cite{chen2022adaptformer}, while extracting task-specific features through task-wise scaling and shift operations (\textbf{V1}$-$\textbf{V4}). The experiments of the VMT-Adapter were conducted by our implementation, as there is no code available. We also implemented our method on the AdaptFormer (\textbf{AO1}$-$\textbf{AO4}) and the VMT-Adapter (\textbf{VO1}$-$\textbf{VO4}). `seq' and `par' indicate the sequential and parallel configurations of the adapter with MLP module as shown in Fig.~\ref{fig:Adapter_config}.
Please refer to the caption of Table~\ref{tab:adapter} for more details on $\rho$ and $r$.

\begin{figure}
  \centering
  \includegraphics[width=0.8\linewidth]{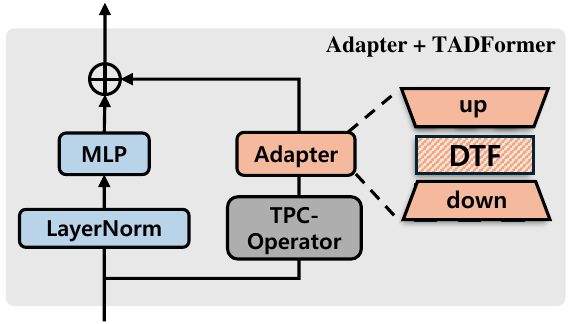} 
  \caption{Overview of Adapter + TADFormer in parallel adapter configuration} 
  \label{fig:Adapter_TADFormer}
\end{figure}

An example of applying our method, TADFormer, to the AdaptFormer~\cite{chen2022adaptformer} is descirbed in Fig.~\ref{fig:Adapter_TADFormer} (\textbf{AO2} or \textbf{AO4}). This design allows the TADFormer module to be integrated into adapter-based PEFT methods, taking into account both task and input contexts. This architecture can be applied to both parallel and sequential configuration of adapters and is equally applicable to other adapter-based method such as VMT-Adapter~\cite{xin2024vmt}. 

In the following, we compared the existing adapter-based methods~\cite{chen2022adaptformer,xin2024vmt} with our method implemented on the adapter framework. For a fair comparison, we evaluated the performance for the cases with the similar amount of trainable parameters, though the results of all possible combinations are provided in Table~\ref{tab:adapter}. Our code on the adapter-based experiments is submitted as supplementary material.

\vspace{5pt}
\noindent\textbf{AdaptFormer vs. AdaptFormer with Ours}.
For a fair comparison in terms of the number of trainable parameters, we compared \textbf{A1, A2} and \textbf{AO3, AO4} that have comparable numbers of trainable parameters.  
In both comparison, TADFormer demonstrated higher $\Delta m$ values when integrated into both sequential and parallel configurations.
This reveals that the application of TADFormer to AdaptFormer results in an overall enhancement in MTL performance.

\vspace{5pt}
\noindent\textbf{VMT-Adapter vs. AdaptFormer with Ours}.
As the VMT-Adapter~\cite{xin2024vmt} is an extension of the AdaptFormer~\cite{chen2022adaptformer}, we applied our method to the AdaptFormer, and then compared it with VMT-Adapter. To be specific, the model proposed in~\cite{xin2024vmt} uses the down-projection ratio of $\rho=4$ (\textbf{V2}). For a fair comparison in terms of the number of trainable parameters, we compared it with the AdaptFormer combined with TADFormer using $r=64$ (\textbf{AO4}). This makes their number of parameters comparable. 
In comparison, \textbf{AO4} achieves a $\Delta m$ increase of 4.07 with only an additional 0.02 M trainable parameters compared to \textbf{V2}. The comparison of \textbf{V1} and \textbf{AO3} also shows a similar tendency. This result demonstrates that the structure of TADFormer is more effective for multi-task learning than the scaling and shift operations used in the VMT-Adapter.



\vspace{5pt}
\noindent\textbf{VMT-Adapter vs. VMT-Adapter with Ours}.
We also compared \textbf{V1}, \textbf{V2} with \textbf{VO3}, \textbf{VO4}, which have comparable numbers of trainable parameters, demonstrating that the performance has notably improved with only 0.3M increase in trainable parameters. This suggests that even when the adapter's down-projection channel dimension is reduced to reduce trainable parameters, the structure of TADFormer is capable of efficiently and effectively extracting multi-task representations. 

To sum up, these results confirm that the TADFormer can be successfully integrated into various adapter-based methods and is the scalable multi-task PEFT method, compatible with both LoRA and adapter-based frameworks.


\begin{table*}[t]
  \centering
  \setlength{\tabcolsep}{8pt} 
  \caption{\textbf{Performance comparison with other decoders:} For the encoder, we use TADFormer ($r=32$) and MTLoRA ($r=32$) with the Swin-T backbone pretrained on ImageNet-22k.} 
  \small
  \resizebox{\textwidth}{!}{ 
  \begin{tabular}{c c c | c c c c | c | c }
    \toprule
     \multicolumn{3}{c|}{\textbf{Model}} & \textbf{SemSeg} & \textbf{Human Parts} & \textbf{Saliency} & \textbf{Normals} & 
      \multirow{2}{*}{$\Delta m (\%) $} & \textbf{Trainable Param.} (M)  \\  
      \textbf{Method} & \textbf{Encoder} & \textbf{Decoder}
      & ($mIoU \uparrow$) & ($mIoU \uparrow$) & ($mIoU \uparrow$) & ($rmse \downarrow$) & &
      Decoder / All \\
      
    \midrule
   \multicolumn{1}{c}{\multirow{3}{*}{\textbf{MTLoRA}~\cite{agiza2024mtlora}}} & \multirow{3}{*}{Swin-T} & HRNet~\cite{sun2019deep} & 69.44 & 61.08 & 63.24 & 16.47 & +2.93 & 1.94 / 6.08 \\
   \multicolumn{1}{c}{} & & SegFormer~\cite{xie2021segformer}  & 69.59 & 61.13 & 63.74 &  16.62 & +3.00 & 2.08 / 6.22  \\ 
   \multicolumn{1}{c}{} & & ASPP~\cite{chen2018encoder} & 72.32 & 60.98 & 63.04 & 16.51 & +3.83 & 12.44 / 16.58  \\  

   \midrule
   \multicolumn{1}{c}{\multirow{3}{*}{\textbf{TADFormer}}} & \multirow{3}{*}{Swin-T} & HRNet~\cite{sun2019deep} & 72.05 & 61.6 & 65.45 & 16.7 & +4.67 & 1.94 / 4.78  \\ 
   \multicolumn{1}{c}{} & & SegFormer~\cite{xie2021segformer} & 72.33 & 61.16 & 65.8 & 16.87 & +4.51 & 2.08 / 4.91  \\  
   \multicolumn{1}{c}{} & & ASPP~\cite{chen2018encoder} & 73.66 & 60.37 & 65.27 & 16.43 & +5.09 & 12.44 / 15.27  \\  

    \bottomrule
  \end{tabular}
  }
  \label{tab:decoder}
\end{table*}

\begin{table*}[t]
  \centering
  \setlength{\tabcolsep}{4.5pt}
\renewcommand{\arraystretch}{0.85}
  \caption{\textbf{Performance comparison with Adapter-based PEFT Methods:} In the adapter-based PEFT, the input feature is down-projected and up-projected within the adapter ($d \rightarrow r \rightarrow$ d), where $d$ is the dimension of an input feature and $r$ is the dimension of hidden layer, which is also called rank. In our experiments, we consider two types of projection dimensions: 1) $\rho=\frac{d}{r}$ denotes the down-projection ratio used in the adapter, 2) $r=64$ denotes a fixed down-projected channel dimension. Additionally, `seq' and `par' indicate the sequential and parallel configurations of the adapter with MLP module as shown in Fig.~\ref{fig:Adapter_config}. 
  This experiment demonstrates the performance of integrating TADFormer with two adapter-based PEFT methods: AdaptFormer and VMT-Adapter. AdaptFormer uses a shared adapter structure for all tasks. VMT-Adapter, similar to AdaptFormer, utilizes a shared adapter but additionally incorporates task-specific scaling and shift operations.
  $\ast$ indicates that the results were reproduced by our implementation, as there is no code available. 
  All results were obtained using the Swin-T pre-trained on ImageNet-1k as in Table~\ref{tab:Quantitive}.}
  
\small
  \begin{tabular}{ c | c | c c c c | c | c }
    \toprule
    \multirow{2}{*}{\textbf{Index}} & \multirow{2}{*}{\textbf{Method}} & \textbf{SemSeg} & \textbf{Human Parts} & \textbf{Saliency} & \textbf{Normals} & 
      \multirow{2}{*}{$\Delta m (\%) $} & \textbf{Trainable}  \\
      & & ($mIoU \uparrow$) & ($mIoU \uparrow$) & ($mIoU \uparrow$) & ($rmse \downarrow$) & &
      \textbf{Parameters} (M) \\
    \midrule
    S & Single Task & 67.21 & 61.93 & 62.35 & 17.97 & 0 & 112.62 \\
    M1 & MTL - Tuning Decoders Only & 65.09 & 53.48 & 57.46 & 20.69 & -9.95  & 1.94 \\
    M2 & MTL - Full Fine Tuning  &  67.56 & 60.24 & 65.21 & 16.64 & +2.23 &  30.06   \\
    \midrule

    A1 & AdaptFormer (seq) \cite{chen2022adaptformer} ($\rho=4$) & 69.01 & 58.2031 & 63.545 & 18.1676 & -0.63 & 3.64 \\
    A2 & AdaptFormer (par) \cite{chen2022adaptformer} ($\rho=4$) & 55.28 & 50.63 & 60.51 & 18.55 & -10.54 & 3.64\\
    A3 & AdaptFormer (seq) \cite{chen2022adaptformer} ($r=64$) & 68.84 & 57.84 & 63.57 & 18.46 & -1.23 & 3.12\\
    A4 & AdaptFormer (par) \cite{chen2022adaptformer} ($r=64$) & 55.18 & 50.39 & 60.36 & 18.78 & -11.06 & 3.12\\

    \midrule
    AO1 & AdaptFormer+Ours (seq) ($\rho=4$) & 72 & 59.62 & 64.94 & 17.4 & 2.69 & 4.48\\ 
    AO2 &AdaptFormer+Ours (par) ($\rho=4$) & 61.41 & 52.93 & 62.88 & 17.57 & -5.02 & 4.48\\ 
 
    AO3 & AdaptFormer+Ours (seq) ($r=64$) & 71.74 & 58.56 & 64.38 & 17.52 & 1.76 & 3.67\\ 
    AO4 & AdaptFormer+Ours (par) ($r=64$) & 60.37 & 52.18 & 62.46 & 17.83 & -6.25 & 3.67\\ 
    
    \midrule
    
    V1 & VMT-Adapter (seq)$^\ast$ \cite{xin2024vmt} ($\rho=4$) & 68.98 & 58.44 & 63.43 & 18.26 & -0.71 & 3.65 \\
    V2 & VMT-Adapter (par)$^\ast$ \cite{xin2024vmt} ($\rho=4$) & 55.4 & 50.98 & 60.45 & 18.5 & -10.32 & 3.65 \\

    V3 & VMT-Adapter (seq)$^\ast$ \cite{xin2024vmt} ($r=64$) & 68.8 & 58 & 63.59 & 18.42 & -1.12 & 3.14 \\ 
    V4 & VMT-Adapter (par)$^\ast$ \cite{xin2024vmt} ($r=64$) & 55.25 & 50.32 & 60.38 & 18.76 & -11.03 & 3.14 \\ 
    
    \midrule
    
    VO1 & VMT-Adapter+Ours (seq) ($\rho=4$) & 71.91 & 59.6 & 64.7 & 17.37 & +2.59 & 4.49\\
    VO2 & VMT-Adapter+Ours (par) ($\rho=4$) & 60.89 & 52.59 & 62.58 & 17.57 & -5.48 & 4.49\\

    VO3 & VMT-Adapter+Ours (seq) ($r=64$) & 71.7 & 58.72 & 64.64 & 17.57 & +1.85 & 3.68\\
    VO4 & VMT-Adapter+Ours (par) ($r=64$) & 59.81 & 51.55 & 62.3 & 17.9 & -6.87 & 3.68\\ 
    
    \midrule
    \bottomrule
  \end{tabular}   
  
  \label{tab:adapter}
\end{table*}

{
    \small
    \bibliographystyle{ieeenat_fullname}
    \bibliography{ref}
}


\end{document}